\documentclass[final,12pt]{clear2024} 

\title[Evaluating and Correcting Performative Effects of Decision Support Systems]{Evaluating and Correcting Performative Effects of Decision Support Systems via Causal Domain Shift}
\usepackage{times}
\usepackage{presets_philip}
\definecolor{cbRed}{HTML}{D55E00}
\definecolor{cbGreen}{HTML}{009E73}

\clearauthor{%
	\Name{Philip Boeken} \Email{p.a.boeken@uva.nl}\\
	\addr University of Amsterdam\\
	\addr Booking.com
	\AND
	\Name{Onno Zoeter} \Email{onno.zoeter@booking.com}\\
	\addr Booking.com
	\AND
	\Name{Joris M. Mooij} \Email{j.m.mooij@uva.nl} \\
	\addr University of Amsterdam%
}

\begin{document}

\maketitle

\begin{abstract}
	When predicting a target variable $Y$ from features $X$, the prediction $\hat{Y}$ can be \emph{performative}: an agent might act on this prediction, affecting the value of $Y$ that we eventually observe. Performative predictions are deliberately prevalent in algorithmic decision support, where a Decision Support System (DSS) provides a prediction for an agent to affect the value of the target variable. When deploying a DSS in high-stakes settings (e.g.\ healthcare, law, predictive policing, or child welfare screening) it is imperative to carefully assess the performative effects of the DSS. In the case that the DSS serves as an alarm for a predicted negative outcome, naive retraining of the prediction model is bound to result in a model that underestimates the risk, due to effective workings of the previous model. In this work, we propose to model the deployment of a DSS as causal domain shift and provide novel cross-domain identification results for the conditional expectation $\EE[Y \given X]$, allowing for pre-~and post-hoc assessment of the deployment of the DSS, and for retraining of a model that assesses the risk under a baseline policy where the DSS is not deployed.
	Using a running example, we empirically show that a \emph{repeated regression} procedure provides a practical framework for estimating these quantities, even when the data is affected by sample selection bias and selective labelling, offering for a practical, unified solution for multiple forms of target variable bias.
\end{abstract}



\section{Introduction}\label{sec:introduction}
When the value of some variable is predicted, this prediction can cause an agent to take action. In the context of linguistics, \cite{austin1962how} coined the term \emph{performative} for utterances that aim at instigating action; in contrast with sentences of a \emph{descriptive} nature. In economics, the concept of performativity has received much attention, and has seen multiple different manifestations. For example, it has been described as a more general concept where the emergence of economic theories legitimize the markets they describe, which caused these markets to become more active. A very concrete type of performativity has been observed in the common use of the Black-Scholes-Merton (BSM) formula for predicting option prices, which in turn affects the price of said options to be close to their predicted value \citep{mackenzie2007do}. Related notions are that of a self-fulfilling prophecy, like the BSM formula, and the self-defeating prophecy, like warning of excessive risk that instigates action to reduce this risk.

In recent machine learning literature, much attention has been given to the performative effects of predictions (\citealp{perdomo2020performative,mendler-dunner2020stochastic,miller2021outside,pombal2022prisoners,kim2023making,yan2023discovering}, among others). A common goal of these works is to make a prediction that is close to the value that will be observed, taking into account the effect that the prediction has on this target variable. Here, a core concept is the minimization of the \emph{performative risk}: the risk of a prediction model, evaluated on the data distribution it entails.

In this work we place the problem of performative prediction in the light of human-algorithmic decision making, where predictions are deliberatively of a performative nature, but do not necessarily have to be close to the eventually observed target variable. For example, algorithms that warn of excessive risk (e.g.\ in churn prediction, predictive policing, or patient monitoring in the ICU) aim at instigating an action that will reduce the predicted risk and thus aim at invalidating the prediction that they make. Such models can be considered to predict risk under the baseline policy where the decision support system (DSS) is not deployed \citep{coston2020counterfactual}. Naive retraining of such prediction models can suffer from a bias that is induced in the training data by the previous prediction algorithm, a concept that we refer to as \emph{performative bias}. In this work, we show how to correct for performative bias by explicitly modelling the deployment of the DSS, and treating the estimation of the \emph{baseline predictor} as a domain adaptation problem.

In aforementioned high-stakes environments, proper evaluation of the DSS is crucial. Over the years many decision support systems have been deployed in high stakes environments, but not all to great success \citep{coston2023validity}. These events motivate thorough testing of any DSS prior to deployment and thorough examination of the system during deployment, to foresee and monitor any undesirable performative effects of the DSS. Despite the urgency of proper continuous assessment of decision support systems, \cite{wu2021how} show that among all medical AI devices that are approved by the FDA between January 2015 and December 2020, most evaluations of those devices are pre-deployment studies, and hardly any post-deployment evaluations have been performed.\footnote{Although not \emph{all} medical AI devices that are considered by \cite{wu2021how} provide explicit decision support, many can be interpreted to do so. For example, image classification techniques for detecting tumours can be seen as providing decision support, and evaluation of the performative effects of the deployment of such AI devices is likely of importance.}
To address the need of evaluation of DSSes, we propose and investigate the estimability of the \emph{deployment effect}, i.e.\ the effect of the deployment of the DSS on the target variable, and of the \emph{retraining effect}, i.e.\ the effect of a new prediction model on the target variable, compared to the average outcome under the previous prediction model. In practise it can be unfeasible or unethical to perform randomized controlled trials with the deployment of a DSS, which makes the estimability of these evaluation metrics a domain adaptation problem.

In the following numerical example we further demonstrate the manifestation of performative bias after naive retraining of a prediction model, and with it the need of its evaluation, e.g.\ by analysis of the retraining effect. This example is inspired by a real-world scenario where in the training data, high-risk individuals receive a treatment that effectively lowers the risk of a negative outcome, inducing a bias in the training data \citep{caruana2015intelligible}.

\begin{figure}[!htb]
	\centering
	\subfigure[$t=1$, DSS off \textcolor{white}{$\hat{Y}$}\label{fig:ex1_naive:ep1}]{
		\centering
		\begin{tikzpicture}
			\node[inner sep=0pt] (b) at (0, 0) {\includegraphics[height=110pt]{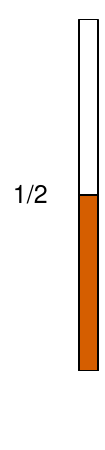}};
			\node[inner sep=0pt] (b) at (2.3, 0) {\includegraphics[height=110pt]{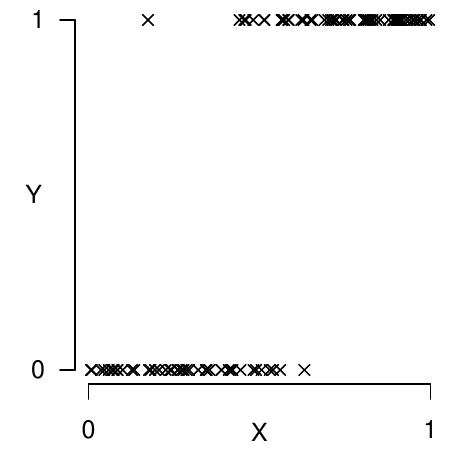}};
			\node[inner sep=0pt] (b) at (0.3, -1.73) {\scriptsize$\EE_1[Y]$};
			\node[inner sep=0pt] (b) at (4.35, -2) {};
		\end{tikzpicture}
	}
	\subfigure[$t=2$, $\hat{Y}=\EE_1(Y \given X)$ \label{fig:ex1_naive:ep2}]{
		\centering
		\begin{tikzpicture}
			\node[inner sep=0pt] (b) at (0, 0) {\includegraphics[height=110pt]{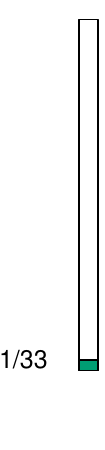}};
			\node[inner sep=0pt] (b) at (2.3, 0) {\includegraphics[height=110pt]{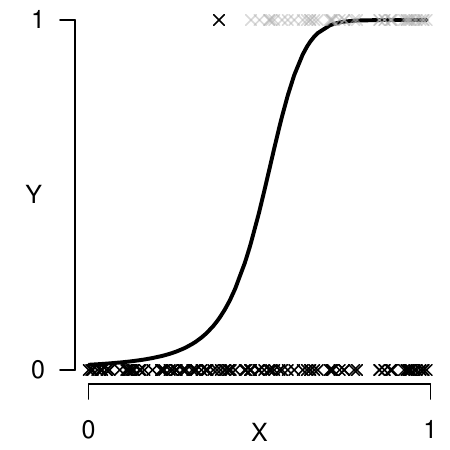}};
			\node[inner sep=0pt] (b) at (0.3, -1.73) {\scriptsize$\EE_2[Y]$};
			\node[inner sep=0pt] (b) at (2.1, -.6) {\scriptsize$\hat{Y}$};
			\draw[arr,draw=cbGreen] (3.3, 1.6) -- (3.3, -1.1);
			\draw[arr,draw=cbGreen] (3.7, 1.6) -- (3.7, -1.1);
			\node[inner sep=0pt] (b) at (4.35, -2) {};
		\end{tikzpicture}
	}
	\subfigure[$t=3, \hat{Y}=\EE_2(Y \given X)$ \label{fig:ex1_naive:ep3}]{
		\centering
		\begin{tikzpicture}
			\node[inner sep=0pt] (b) at (0, 0) {\includegraphics[height=110pt]{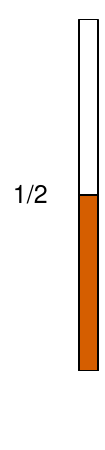}};
			\node[inner sep=0pt] (b) at (2.3, 0) {\includegraphics[height=110pt]{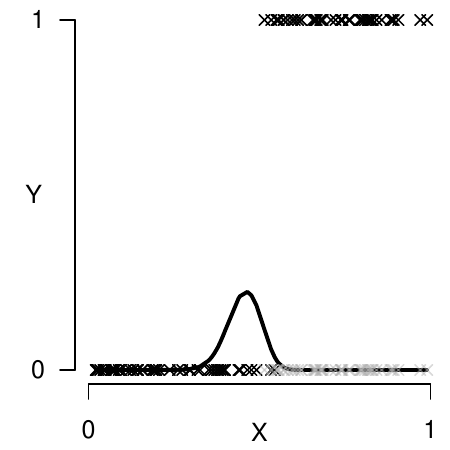}};
			\node[inner sep=0pt] (b) at (0.3, -1.73) {\scriptsize$\EE_3[Y]$};
			\node[inner sep=0pt] (b) at (2.1, -.6) {\scriptsize$\hat{Y}$};
			\draw[arr,draw=cbRed] (3.3, -1.1) -- (3.3, 1.6);
			\draw[arr,draw=cbRed] (3.7, -1.1) -- (3.7, 1.6);
			\node[inner sep=0pt] (b) at (4.35, -2) {};
		\end{tikzpicture}
	}
	\caption{In epoch $t=1$ the DSS is not deployed. In $t=2$ a DSS $\hat{Y}$ is deployed that is trained on data from $t=1$, effectively reducing the mean of $Y$. In $t=3$, a DSS $\hat{Y}$ that is naively retrained on data from $t=2$ is deployed, increasing the mean of $Y$.}
	\label{fig:ex1_naive}
\end{figure}
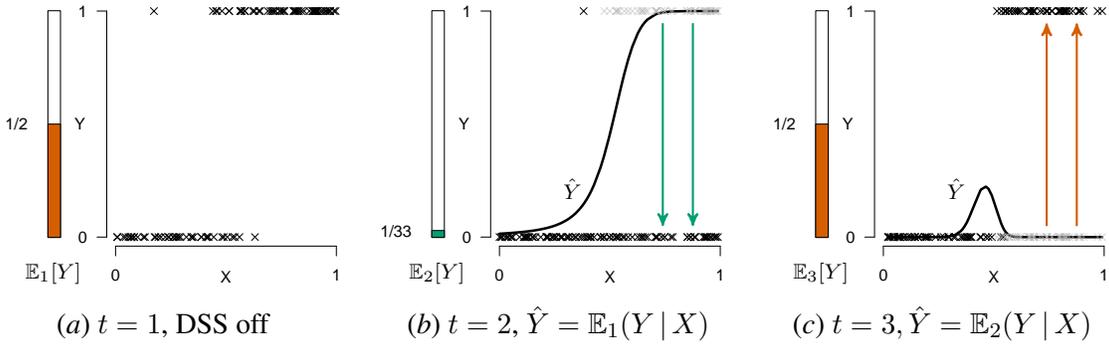

\begin{example}\label{ex:naive_fails}
	Let $X\sim \mathrm{Unif}[0,1], \hat{Y} = f(X)$ for some function $f$, and $Y \sim \mathrm{Ber}(\sigma(X-1/2)\I\{\hat{Y} < 1/2\})$ with $\sigma(x) = (1+e^{-x})^{-1}$. Three `epochs' (samples) of this data generating process are shown in Figure \ref{fig:ex1_naive}. In the first epoch, the DSS is not deployed, so we let $\hat{Y} \equiv 0$. In the second epoch, a DSS that is trained on data from epoch one is deployed, so we let $\hat{Y} = \hat{\EE}_1[Y|X]$,\footnote{We let $\hat{\EE}[Y \given X]$ denote an estimate of the conditional expectation $\EE[Y \given X]$. This should interchangeably be interpreted as the function $x\mapsto \hat{\EE}[Y \given X=x]$ or as the evaluation $\hat{\EE}[Y \given X=X]$.} estimated from $\PP_1(X, Y)$. To units where the predicted risk $\hat{Y}$ exceeds the threshold $1/2$, action is taken to greatly reduce this risk, effectively setting $\PP(Y=1 \given X, \{\hat{Y} > 1/2\}) = 0$ and thus preventing the outcome $Y=1$. The green arrows signify this positive effect, marking the grey-coloured counterfactual observations that would have had the value $Y=1$ if no prediction had been made, and which have the value $Y=0$ now that the prediction is made.
	This improvement is also indicated by the bar charts, showing $\EE_2[Y] \approx 1/33 < 1/2 \approx \EE_1[Y]$. In the third epoch, a DSS is naively trained on data from the second epoch, resulting in a model that underestimates the risk, due to effective workings of the DSS in the previous epoch. Hence, the re-training has a negative effect on the average outcome, as indicated by the red arrows and by the bar chart for $\EE_3[Y]$.
\end{example}

\paragraph{Contributions}
In this work, we model the deployment of decision support systems as causal domain shift, and we investigate two applications of this causal model. The first application is the evaluation of whether a novel DSS should be deployed, an existing DSS should be taken offline, or whether a retrained version should be deployed. We define the \emph{deployment effect} and \emph{retraining effect} as suitable evaluation metrics, and we show that the estimation of these evaluation metrics constitutes two domain adaptation tasks (T1 and T2). The second application concerns the estimation of a prediction model to be used by the DSS. We show that naive retraining of such prediction models gets affected by \emph{performative bias} that is induced by the previous prediction model; correcting for this bias constitutes another domain adaptation problem (T3). We show that these domain adaptation tasks are mathematically equivalent, and that they are not solvable (without additional assumptions besides the causal model) when one cannot perform randomized experiments with the deployment of the DSS. We define a \emph{domain pivot} as a set of variables that, when measured in both the source- and target domain of the domain adaptation problem, provides a solution to the domain adaptation problems T1--3, and hence to the evaluation and bias correction applications. We employ the \emph{repeated regression} estimator from \cite{boeken2023correcting} for estimating the quantities of interest. As this estimator has originally been devised to deal with selection bias, we generalise the identifiability and estimation results to settings that are subject to selection bias and/or selective labelling (missing response). Efficacy of these methods is subsequently shown using Example \ref{ex:naive_fails}.

\subsection{Related work}
A line of work following from \cite{perdomo2020performative} considers the general setting where model parameters $\theta$ for making a prediction $\hat{\EE}_\theta[Y \given X]$ induce a shift of the distribution of $(X, Y)$. This dependence can be made explicit by writing $\PP(X, Y \given \theta)$. Similar to \cite{mendler-dunner2022anticipating} and \cite{kim2023making}, we consider the specific setting of \emph{outcome performativity} where the parameters don't affect the features $X$ but only the outcome $Y$, so where the distribution factorizes according to $\PP(X, Y \given \theta) = \PP(Y \given X, \theta)\PP(X)$, and conditional on the parameters the $(X, Y)$ pairs are drawn i.i.d. This is a different setup than e.g.\ \cite{chen2023model} consider, as they allow effects like $\hat{Y}_i \to X_j$ and $\hat{Y}_i \to Y_j$, where $i\neq j$ are sample indicators. We extend the setting of \cite{mendler-dunner2022anticipating} and \cite{kim2023making} by explicitly considering the domain where the DSS is not deployed, allowing for the formulation of the domain adaptation task that we consider. For more details we refer to Appendix \ref{app:perf_pred}.

The task of transporting a statistical relation $\EE[Y \given X]$ over such domains is considered in the line of work on \emph{transportability}. Similar to \cite{pearl2011transportability} and \cite{magliacane2018domain} we leverage sets of variables that render a target variable independent from a domain indicator (which we refer to as \emph{domain pivots}) to transfer statistical relations over domains. Sound and complete algorithms for transporting statistical relations are for example given by \cite{correa2019statistical} and \cite{lee2020general}. However, these algorithms make weaker assumptions than we do (in terms of available data), which makes them unable to identify the target quantities that we consider.

The work of \cite{coston2020counterfactual} considers risk estimation under binary treatment, similar to how we estimate the effect of deployment on an outcome variable $Y$. However, their estimation method requires stronger assumptions on the available data than we do, making it unsuitable for the setting that we consider.

In special cases where the utilized \emph{domain pivot} consists of a context $X$ and an \emph{action variable} (with a finite state space) that the agent controls to optimize a reward $Y$, our proposed evaluation method can be interpreted as a form of off-policy evaluation for contextual bandits, as investigated by \cite{dudik2014doubly,wang2017optimal}. We elaborate on this connection in Appendix \ref{app:off-policy}.

To estimate our quantities of interest, we employ the \emph{repeated regression} estimator from \cite{boeken2023correcting}. This estimator is originally proposed to correct for selection bias. In this work, we show that this estimator can simultaneously correct for selection bias \emph{and} performative bias. The repeated regression estimator bears resemblance to the work on surrogate indices by \cite{athey2019surrogate}, as both methods consider the use of a conditional expectations as pseudo-labels in the estimation procedure. However, translating it to our setting, the work on surrogate indices operates under a different set of assumptions than we do, as it requires the target variable $Y$ to be measured under both deployment and non-deployment. More details are provided in Appendix \ref{app:surrogates}.

\section{Causal modelling of decision support systems}\label{sec:domain_adaptation}
We consider the setting with multidimensional covariates $X$, univariate target variable $Y$, and a prediction $\hat{Y}$ of $Y$ that is a function of $X$ and parameters $\Theta$, denoted with $\hat{Y} = f_{\hat{Y}}(X, \Theta)$.
We allow for hidden confounding between $X$ and $Y$. We assume the state space of any variable $V$ to be (a subset of) $\RR^{d_V}$ for some $d_V \in \NN$, equipped with its standard topology and the Borel sigma-algebra. Variables, their state space, and their values are indicated with uppercase, calligraphic and lowercase letters ($V, \Vcal, v$) respectively.

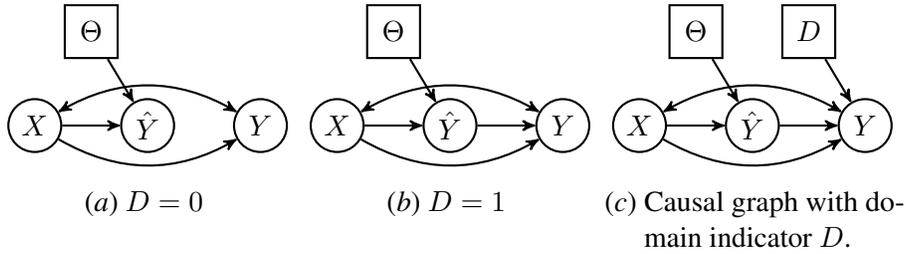
\begin{figure}[ht]
	\centering
	\subfigure[$D=0$ \label{fig:contextual_graph_d0}]{
		\centering
		\begin{tikzpicture}
			\node[var] (X) at (0, 0) {$X$};
			\node[var] (Yh) at (1.5, 0) {$\hat{Y}$};
			\node[var] (Y) at (3, 0) {$Y$};
			\node[vari] (theta) at (0.75, 1.25) {$\Theta$};
			\draw[arr] (X) to (Yh);
			\draw[arr, bend right] (X) to (Y);
			\draw[biarr, bend left] (X) to (Y);
			\draw[arr] (theta) to (Yh);
		\end{tikzpicture}
	}
	\subfigure[$D=1$ \label{fig:contextual_graph_d1}]{
		\centering
		\begin{tikzpicture}
			\node[var] (X) at (0, 0) {$X$};
			\node[var] (Yh) at (1.5, 0) {$\hat{Y}$};
			\node[var] (Y) at (3, 0) {$Y$};
			\node[vari] (theta) at (0.75, 1.25) {$\Theta$};
			\draw[arr] (X) to (Yh);
			\draw[arr, bend right] (X) to (Y);
			\draw[arr] (Yh) to (Y);
			\draw[biarr, bend left] (X) to (Y);
			\draw[arr] (theta) to (Yh);
		\end{tikzpicture}
	}
	\subfigure[Causal graph with domain indicator $D$. \label{fig:contextual_graph}]{
		\centering
		\begin{tikzpicture}
			\node[var] (X) at (0, 0) {$X$};
			\node[var] (Yh) at (1.5, 0) {$\hat{Y}$};
			\node[var] (Y) at (3, 0) {$Y$};
			\node[vari] (theta) at (0.75, 1.25) {$\Theta$};
			\node[vari] (D) at (2.25, 1.25) {$D$};
			\draw[arr] (X) to (Yh);
			\draw[arr, bend right] (X) to (Y);
			\draw[arr] (Yh) to (Y);
			\draw[biarr, bend left] (X) to (Y);
			\draw[arr] (theta) to (Yh);
			\draw[arr] (D) to (Y);
		\end{tikzpicture}
	}
	\caption{Modelling the deployment of the DSS with prediction $\hat{Y}$ as domain shift.}
	\label{fig:contextual_graphs}
\end{figure}

To distinguish between pre- and post-deployment settings of the decision support system, we introduce a domain indicator $D$ which represents a context-specific dependency: $D=0$ indicates the domain where $\hat{Y}$ does not affect $Y$ (i.e.\ the prediction is not published), and $D=1$ indicates the domain where $\hat{Y}$ affects $Y$. Formally, the structural causal model\footnote{We will use many concepts from this causal framework: parents, children, ancestors, d-separation, the Markov property, etc. For more information, we refer to \cite{pearl2009causality} and \cite{bongers2021foundations}.} to describe this data generating process is
\begin{equation}
	X = f_X(E_X, E_{XY}), \quad \hat{Y} = f_{\hat{Y}}(X, \Theta), \quad Y = \begin{cases}
		f_{Y,0}(X, E_{XY}, E_Y)          & \text{if $D=0$} \\
		f_{Y,1}(X, \hat{Y}, E_{XY}, E_Y) & \text{if $D=1$}
	\end{cases}
\end{equation}
with independent exogenous distributions $\PP(E_X), \PP(E_Y)$ and $\PP(E_{XY})$ and measurable functions $f_X, f_{\hat{Y}}, f_{Y,0}, f_{Y,1}$. The Acyclic Directed Mixed Graph (ADMG) of this SCM in the different domains $D$ is depicted in Figures \ref{fig:contextual_graph_d0} and \ref{fig:contextual_graph_d1}, and the causal graph of the joint model is depicted in Figure \ref{fig:contextual_graph}, with explicit domain indicator $D$.

As we don't assume an a-priori distribution for the variables $D$ and $\Theta$, these variables are graphically indicated with squared nodes and formally referred to as \emph{input nodes} of the SCM, following \cite{forre2021transitional}. For given values of $D$ and $\Theta$, this SCM gives rise to the Markov kernel $\PP(X, \hat{Y}, Y \given \Do(D, \Theta))$, defined as the pushforward of the exogenous distribution $\PP(E_X)\otimes\PP(E_Y)\otimes\PP(E_{XY})$ through the structural equations.\footnote{We define $\PP(X, Y \given \Do(D=0))$ (without dependence on $\Theta$) similarly but with the structural equation for $Y$ evaluated at $D=0$, for which we have $\PP(X, Y \given \Do(D=0)) = \PP(X, Y \given \Do(D=0, \Theta=\theta))$ for all $\theta$.} We assume that data will be sampled over epochs, indicated by $t$, where for given values $\theta_t, d_t$ we sample $(X_{t,i}, \hat{Y}_{t,i}, Y_{t,i}) \sim \PP(X, \hat{Y}, Y \given \Do(D = d_t, \Theta=\theta_t))$ i.i.d.\ for $i=1,...,n_t$ and some $n_t\in\NN$. Note that this implies that $D$ and $\Theta$ are neither influenced by, nor confounded with the variables $(X_{t,i}, \hat{Y}_{t,i}, Y_{t,i})$ of the current epoch. Denoting measurements $(V_{t,1}, ...,V_{t,n_t})^T$ of a variable $V$ with $\bm{V}_t$, the values $\theta_t$ and $d_t$ can be determined by data from past epochs $\{(\bm{X}_s, \bm{\hat{Y}}_s, \bm{Y}_s, \theta_s, d_s): s < t\}$.

When considering SCMs with more variables than $\{X, \hat{Y}, Y, D, \Theta\}$, we require the latent projection onto $\{X, \hat{Y}, Y, D, \Theta\}$ to be (a subgraph of) the graph from Figure \ref{fig:contextual_graph} for it to appropriately represent the deployment of a DSS.
\begin{assumption}\label{ass:pp_graph}
	We consider the set $\Mcal$ of SCMs with endogenous variables $V\supseteq\{X, \hat{Y}, Y\}$, input variables $\{D, \Theta\}$ and graph $G$ such that $\Pa_G(\hat{Y}) = \{X, \Theta\}$ and $\Ch_G(D) = \Ch_G(\hat{Y})$, and such that the graph of the latent projection of $G$ onto $\{X, \hat{Y}, Y, D, \Theta\}$ is a subgraph of the ADMG in Figure \ref{fig:contextual_graph}.
\end{assumption}

As alluded to in Section \ref{sec:introduction}, we are interested in the evaluation of the DSS prior to- and during its deployment, and in correcting for the bias that is induced by a previous deployment of the DSS when retraining the prediction model. Having explicitly defined the deployment indicator $D$ and parameters $\Theta$, we are enabled to make these estimation tasks more precise.

\subsection{Application A: Evaluation}
When a DSS with some parameter value $\theta$ and prediction model $\hat{Y} = f_{\hat{Y}}(X, \theta)$ has been developed, the intention behind this DSS is to improve the value of some outcome metric $Y$. As motivated in the introduction, human usage of a newly developed DSS can involve errors that have a negative impact on $Y$. To evaluate this, we define the \emph{deployment effect} of a DSS with parameters $\theta$.

\begin{definition}[Deployment effect]
	The \emph{deployment effect} of a DSS with parameters $\theta$ is defined as the average causal effect of the deployment of the DSS on the target variable, i.e.
	\begin{equation}
		\tau(\theta) := \EE[Y|\Do(D=1, \Theta = \theta)] - \EE[Y \given \Do(D=0)].
	\end{equation}
\end{definition}
Prior to deployment of the DSS we are interested in estimating $\tau(\theta)$ from data sampled from $\PP(X, Y \given \Do(D=0))$. Since $\EE[Y|\Do(D=0)]$ is directly estimable, the challenge lies in estimating $\EE[Y|\Do(D=1, \Theta=\theta)]$. We refer to this domain adaptation task as \textbf{T1.a}.
After deployment, we are interested in estimating $\tau(\theta)$ from $\PP(X, Y \given \Do(D=1, \Theta=\theta))$, e.g.\ to monitor the correct usage of the DSS: if the mean value of $Y$ is estimated to be worse for $(D=1, \Theta=\theta)$ than for $D=0$, it could be better to turn off the DSS, and further investigate why it has a negative effect on the outcome. Since $\EE[Y|\Do(D=1,\Theta=\theta)]$ is directly estimable, the challenge then lies in estimating $\EE[Y|\Do(D=0)]$. We refer to this domain adaptation task as \textbf{T1.b}.

\medskip
When in epoch $t$ a DSS with parameters $\theta_t$ is deployed one might further develop the DSS, resulting in parameters $\theta_{t+1}$. Before deploying this `retrained' DSS, one might want to evaluate the impact that these new parameters will have on $Y$.
\begin{definition}[Retraining effect]
	The \emph{retraining effect} is defined as the average causal effect of the deployment of a retrained DSS on the target variable, i.e.
	\begin{equation}
		\rho(\theta_{t+1}, \theta_{t}) := \EE[Y|\Do(D=1, \Theta = \theta_{t+1})] - \EE[Y \given \Do(D=1, \Theta = \theta_t)].
	\end{equation}
\end{definition}
In the setting described above, we aim at estimating $\rho(\theta_{t+1}, \theta_t)$ from $\PP(X, Y \given \Do(D=1, \Theta=\theta_t))$. Since $\EE[Y|\Do(D=1, \Theta=\theta_t)]$ is directly estimable, the challenge lies in estimating $\EE[Y|\Do(D=1, \Theta=\theta_{t+1})]$. We refer to this domain adaptation task as \textbf{T2}.
We have summarised these domain adaptation tasks in Table \ref{tab:da_tasks}.

\begin{table}[tb]
	\centering
	\begin{tabular}{lllll}\toprule
		              & \textbf{Metric}                & \textbf{Source domain}   & \textbf{Target domain}     & \textbf{Target quantity}                      \\
		\midrule
		\textbf{T1.a} & $\tau(\theta)$                 & $D=0$                    & $D=1, \Theta=\theta$       & $\EE[Y \given \Do(D=1, \Theta=\theta)]$       \\
		\textbf{T1.b} & $\tau(\theta)$                 & $D=1, \Theta=\theta$     & $D=0$                      & $\EE[Y \given \Do(D=0)]$                      \\
		\textbf{T2}   & $\rho(\theta_{t+1}, \theta_t)$ & $D=1, \Theta=\theta_{t}$ & $D=1, \Theta=\theta_{t+1}$ & $\EE[Y \given \Do(D=1, \Theta=\theta_{t+1})]$ \\
		\textbf{T3}   & --                             & $D=1, \Theta=\theta$     & $D=0$                      & $\EE[Y \given X, \Do(D=0)]$                   \\
		\bottomrule
	\end{tabular}
	\caption{Domain adaptation tasks for evaluation (T1, T2) and performative bias correction (T3).}
	\label{tab:da_tasks}
\end{table}

\subsection{Application B: Bias correction}
Let $Y$ be an outcome whose expected value we want to minimize, e.g.\ a cost, negative utility, or negative reward. Prior to deployment, data is generated from $\PP(X, Y \given \Do(D=0))$, and the average outcome $Y$ is related to features $X$ via $\EE[Y \given X, \Do(D=0)]$. This could be considered to be a `baseline policy'. It might be the case that a DSS is developed to identify cases that have (under this baseline policy) a high risk of seeing an outcome that is to be prevented, like a patients death, a customer churning, or a crime to be committed in a particular neighbourhood \citep{coston2020counterfactual}. In this setting, a sensible predictor $\hat{Y}$ would be the following:
\begin{definition}[Baseline predictor]
	We are interested in estimating the \emph{baseline predictor}
	\begin{equation}
		\hat{Y}_{bp} = \EE[Y \given X, \Do(D=0)].
	\end{equation}
\end{definition}
As demonstrated in Example \ref{ex:naive_fails}, naive regression of $Y$ on $X$ to retrain the model for $\hat{Y}$ from $\PP(X, Y \given \Do(D=1, \Theta))$ would yield a predictor $\hat{Y} = \EE[Y \given X, \Do(D=1, \Theta)]$, and hence is biased when the DSS is supposed to make the baseline prediction.
\begin{definition}[Performative bias]
	The \emph{performative bias} is defined as the bias that the deployment of the DSS induces on the statistical relation $\EE[Y \given X]$, i.e.
	\begin{equation}
		\EE[Y \given X, \Do(D = 1, \Theta=\theta)] - \EE[Y |X, \Do(D=0)].
	\end{equation}
\end{definition}
Estimating the baseline predictor from the domain $(D=1, \Theta=\theta)$, and thus correcting for performative bias, is a domain adaptation task that we refer to as \textbf{T3}.

\medskip
If we let $Y$ be binary, the baseline predictor is indeed the optimal prediction function $\hat{Y} : \Xcal \to [0,1]$ if it is a risk assessment for the event $Y=1$ (given features $X$) and serves as an `alarm' to identify `risky cases', based on which an agent can take an action $A$ which surely decreases the risk to a known level, but which one also wants to use sparingly. The action $A$ could for example be to operate a patient with features $X$ to minimize the probability of death $Y$, or the offering of a discount $A$ to a customer $X$ to minimize the probability of churning $Y$. Clearly, one wants to use these actions sparingly. This type of optimality of the baseline predictor is formalised as follows:

\begin{proposition}\label{thm:baseline_predictor_optimal}
	Given a Markov kernel $\PP(Y=1 | X, A)$, consider the SCM $X\sim \PP(X), A = D \cdot \I{\{\hat{Y} > \varepsilon(X)\}}, \hat{Y} = \hat{y}(X), Y \sim \PP(Y=1 \given X, A)$ with $\varepsilon(x) := \PP(Y=1 \given X=x, A=1)$ and some function $\hat{y}:\Xcal \to [0,1]$. The baseline predictor $\hat{Y}_{bp}$ solves the following bilevel optimization problem:
	\begin{align}
		\begin{split}
			1. & ~~ H := \argmin_{\hat{y} \in [0,1]^\Xcal}\PP(Y=1 \given X=x, \Do(D=1, \hat{Y} = \hat{y}(x))) \\
			2. & ~~ \hat{Y}_{bp} \in \argmin_{\hat{y}\in H} \PP(A=1 \given X=x, \Do(D=1, \hat{Y} = \hat{y}(x)))
		\end{split}
	\end{align}
	for $\PP(X)$-almost all $x\in \Xcal$.
\end{proposition}

\subsection{Equivalence of T1--3 and their non-identifiability}
Having these two applications in mind, our goal is to estimate the deployment effect $\tau(\theta)$, the retraining effect $\rho(\theta_{t+1}, \theta_t)$, and the baseline predictor $\hat{Y}_{bp}$ from varying source domains, and thus solving domain adaptation tasks T1--3 as displayed in Table \ref{tab:da_tasks}. A prerequisite for estimation is the \emph{identifiability} of these quantities: whether there exists a mathematical operation on the source distribution that yields the target quantity. This concept is formally defined as follows:
\begin{definition}[Identifiability]
	Given a set of SCMs $\Mcal$, a target quantity $t(M)$ (some function of $M\in\Mcal$) is \emph{identifiable} in $\Mcal$ from a set $s(M) := \{s_1(M), ..., s_n(M)\}$ of (marginal, conditional and/or interventional) distributions induced by $M$ if $s(M_1) = s(M_2) \implies t(M_1) = t(M_2)$ for all $M_1, M_2 \in \Mcal$.
\end{definition}

Throughout, we will consider $\Mcal$ as defined in Assumption \ref{ass:pp_graph}. Given source distribution(s) $s(M)$ and target quantity $t(M)$, identifiability means that for all $M\in\Mcal$ the map $s(M) \mapsto t(M)$ is well defined, which can be shown by explicitly providing it.
Non-identifiability can be shown by providing SCMs $M_1, M_2\in\Mcal$ for which $s(M_1) = s(M_2)$ but $t(M_1) \neq t(M_2)$.
Throughout, our target $t(M)$ will be a conditional expectation, and we will specify $s(M)$ and $t(M)$ without explicit dependence on $M$.

We first consider the graph depicted in Figure \ref{fig:contextual_graph} and we show that tasks T1--3 are equivalent to a single domain adaptation task:
\begin{lemma}\label{thm:da_reduces_to_regression}
	Identifiability of each of the target quantities of the domain adaptation tasks T1--3 is equivalent to identifiability of the conditional expectation $\EE[Y \given X, \Do(D = d, \Theta = \theta)]$ from $\PP(X, Y | \Do(D=d', \Theta=\theta'))$ for some $(d', \theta') \neq (d, \theta)$ with $d' = d = 1 \iff \theta'\neq\theta$.
\end{lemma}
The last condition on the domains ensures that we don't consider $d=d'=0$ and $\theta \neq \theta'$, which is trivially excluded as the distribution of $(X, Y)$ would then be the same in the source and target domains.

The following proposition shows that measuring a single source distribution is not sufficient for identification of the target quantities as specified above, and with that, that the tasks T1--3 cannot be solved without imposing additional assumptions.
\begin{proposition}\label{thm:unbiased_ce_not_identified}
	The target quantity $\EE[Y \given X, \Do(D=d, \Theta=\theta)]$ is not identifiable in $\Mcal$ from $\PP(X, Y \given \Do(D=d', \Theta=\theta'))$ if $(d',\theta') \neq (d,\theta)$ with $d' = d = 1 \iff \theta'\neq\theta$.
\end{proposition}

It is immediately clear that if $(X, Y)$ pairs are measured in both the source and target domains, then $\EE[Y|X, \Do(D=d, \Theta=\theta)]$ would be identifiable, and hence also $\tau(\theta), \rho(\theta_{t+1}, \theta_t)$ and $\hat{Y}_{bp}$.
However, in practice it might not be feasible to gather labels from the target domain. Alas, in high-stakes settings, deploying a DSS without knowing what effect it will have on the outcome $Y$ can be undesirable. So, to be able to identify these target quantities without requiring measurements of the outcome variable $Y$ in the target domain we will leverage additional assumptions, as demonstrated in the next section.

\subsection{Domain pivots: mediators of the prediction and outcome}\label{sec:domain_adaptation:domain_pivot}
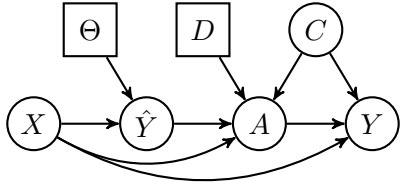
\begin{figure}[ht]
	\centering
	\begin{tikzpicture}
		\node[var] (X) at (0, 0) {$X$};
		\node[var] (Yh) at (1.5, 0) {$\hat{Y}$};
		\node[var] (A) at (3, 0) {$A$};
		\node[var] (Z2) at (3.75, 1.25) {$C$};
		\node[var] (Y) at (4.5, 0) {$Y$};
		\node[vari] (theta) at (0.75, 1.25) {$\Theta$};
		\node[vari] (D) at (2.25, 1.25) {$D$};
		\draw[arr] (X) to (Yh);
		\draw[arr, bend right] (X) to (A);
		\draw[arr, bend right] (X) to (Y);
		\draw[arr] (theta) to (Yh);
		\draw[arr] (Yh) to (A);
		\draw[arr] (D) to (A);
		\draw[arr] (A) to (Y);
		\draw[arr] (Z2) to (A);
		\draw[arr] (Z2) to (Y);
	\end{tikzpicture}
	\caption{Performative prediction through a mediator $A$, with an observed common cause $C$.}
	\label{fig:mediator}
\end{figure}

When a prediction $\hat{Y}$ affects the value $Y$, this might happen through an \emph{action} $A$ that perfectly mediates $\hat{Y}$ and $Y$, i.e.\ we have $\hat{Y}\to A \to Y$ and there is no edge $\hat{Y}\to Y$. In this case, only $A$ is directly affected by the deployment of the DSS, so we have $D\to A$ and not $D\to Y$. If this action $A$ is unconfounded with $Y$, we have the independence $Y\Indep D,\Theta \given X, A$. If $A$ and $Y$ are confounded (by $C$, say), we have $Y\Indep D,\Theta \given X, Z$ where $Z=\{A, C\}$. A graphical depiction of this setting is provided in Figure \ref{fig:mediator}. As we will see later, finding a set of variables $Z$ for which this conditional independence is satisfied is instrumental for the domain adaptation task that we have in mind. Note that Figure \ref{fig:mediator} is not the only graph that exhibits $Y\Indep D,\Theta \given X, C, A$, as we can add multiple instances of latent confounding (bidirected edges) to this graph and maintain the required independence.

\begin{definition}[Domain pivot]
	Given domain indicator $D$, features $X$, target variable $Y$ and estimand $F(\PP(Y|X, D=d))$ with $F$ a statistical functional\footnote{$F$ is a statistical functional if it is a function $F:\Pcal(\Ycal) \to \RR^d$, with $\Pcal(\Ycal)$ the space of probability distributions on $\Ycal$, and $d\in\NN$. For more information, see \cite{shao2003mathematical}.} and $d\in\Dcal$, we call $\{X, Z\}$ a \emph{domain pivot} for $F(\PP(Y|X, D=d))$ if $Y\Indep D \given X, Z$.
\end{definition}

Our main solution for T1--3 assumes that the domain pivot $\{X, Z\}$ can be measured in the target domain. Sampling from $\PP(X, Z \given \Do(D=1, \Theta))$ prior to deployment of the DSS might seem like an unreasonable assumption, but there can be practical ways to do so. Consider the example of a patient with features $X$, and a doctor having to decide treatment $A$, where $\{X, A\}$ is a domain pivot. Here, $\PP(A |X, \hat{Y}, \Do(D=1, \Theta))$ can be measured by showing the doctor the prediction $\hat{Y}$, and measuring the treatment that the doctors prescribe for this patient after seeing this prediction. Measuring such intended actions is also leveraged by \cite{stensrud2023optimal} to improve treatment regimes. Practically, one would require the availability of another doctor who has not seen the prediction of the DSS to prescribe the treatment that will actually be carried out.

Similarly, if a DSS is currently deployed one can sample from $\PP(X, A \given \Do(D=0))$ without taking the DSS offline, by measuring an intended action $A$ without revealing the prediction $\hat{Y}$ to the agent. After having made this measurement, one can reveal the prediction $\hat{Y}$, and the agent can proceed with taking actions.

If $A$ and $Y$ are confounded by a common cause $C$ (so $Y\nIndep D, \Theta \given X, A$ and $Y\Indep D, \Theta \given X, A, C$) then it is instrumental to also measure this confounding information, i.e. measure $\PP(X, A, C| \Do(D=d, \Theta=\theta))$ for target domain $(d, \theta)$. This is a restrictive, but common assumption in causal inference. In automated decision making, it is not uncommon for a decision algorithm to heuristically combine a prediction $\hat{Y}$ and additional covariates $C$ that were not used for making the prediction (so they are not part of features $X$), in which case this confounding information might be readily available.

Note that prior to deployment, one cannot test for the required independence due to absence of labels $Y$ from the domain $D=1$. Instead, one could motivate this independence assumption by causal modelling of the data generating process.

Our main result is that when it is unfeasible to measure labels $Y$ in the target domain, but when we are able to measure variables $\{X, Z\}$ in the target domain, our target quantities can be identified if and only if $\{X, Z\}$ is a domain pivot.

\begin{proposition}\label{thm:id_complete}
	Let $(d',\theta') \neq (d,\theta)$ be given with $d' = d = 1 \iff \theta'\neq\theta$. The target quantity $\EE[Y \given X, \Do(D=d, \Theta=\theta)]$ is identifiable in $\Mcal$ from
	\begin{equation}
		\{\PP(X, Y, Z \given \Do(D=d', \Theta=\theta')), \PP(X, Z \given \Do(D=d, \Theta=\theta))\}
	\end{equation}
	if $\PP(X, Z \given \Do(D=d, \Theta=\theta)) \ll \PP(X, Z \given \Do(D=d', \Theta=\theta'))$,\footnote{For two distributions $\PP(X)$ and $\tilde{\PP}(X)$, $\PP(X) \ll \tilde{\PP}(X)$ denotes absolute continuity of $\PP$ with respect to $\tilde{\PP}$, i.e. $\PP(X \in B) > 0 \implies \tilde{\PP}(X\in B) > 0$ for all measurable sets $B$.} and if and only if $Y\Indep D, \Theta \given X, Z$, in which case
	\begin{equation}\label{eqn:id_unbiased_model}
		\EE[Y \given X, \Do(D=d, \Theta=\theta)] = \int \EE[Y \given X, Z, \Do(D=d', \Theta=\theta')]\diff\PP(Z \given X, \Do(D=d, \Theta=\theta))
	\end{equation}
	$\PP(X|\Do(D=d, \Theta=\theta))$-a.e.
\end{proposition}
The absolute continuity ensures that the conditional expectation $\EE[Y \given X, Z, \Do(D=d', \Theta=\theta')]$ is $\PP(X, Z|\Do(D=d, \Theta=\theta))$-a.e.\ well-defined, and hence that the integral in (\ref{eqn:id_unbiased_model}) is well-defined.

Via Lemma \ref{thm:da_reduces_to_regression} and Proposition \ref{thm:id_complete}, we have solvability of domain adaptation tasks T1--3 under the assumption that a domain pivot is measured in the source and target domain. For completeness, we provide an overview of these implied identifiability results:
\begin{corollary}\label{thm:id_applications}
	In the subset of SCMs of $\Mcal$ that have a domain pivot $\{X, Z\}$ for $\EE[Y|X, \Do(D, \Theta)]$ and for which $\PP(X, Z \given \Do(D=d, \Theta=\theta)) \ll \PP(X, Z \given \Do(D=d', \Theta=\theta'))$ for all $d,d', \theta, \theta'$, we have that
	\begin{enumerate}
		\item[T1.] the deployment effect $\tau(\theta)$ is identifiable from $\{\PP(X, Y, Z \given \Do(D=0)), \PP(X, Z \given \Do(D=1, \Theta=\theta))\}$ via
			\begin{equation}\label{eqn:t1a}
				\tau(\theta) = \EE[\EE[Y \given X, Z, \Do(D=0)] \given \Do(D=1, \Theta=\theta)] - \EE[Y|\Do(D=0)]
			\end{equation}
			and from $\{\PP(X, Y, Z \given \Do(D=1, \Theta=\theta)), \PP(X, Z \given \Do(D=0))\}$ via
			\begin{equation}\label{eqn:t1b}
				\tau(\theta) = \EE[Y \given \Do(D=1, \Theta=\theta)] - \EE[\EE[Y \given X, Z, \Do(D=1, \Theta=\theta)]|\Do(D=0)];
			\end{equation}
		\item[T2.] the retraining effect $\rho(\theta_{t+1}, \theta_t)$ is identifiable from $\{\PP(X, Y, Z \given \Do(D=1, \Theta=\theta_t)), \PP(X, \\ Z \given \Do(D=1, \Theta=\theta_{t+1}))\}$ via
			\begin{multline}\label{eqn:t2}
				\rho(\theta_{t+1}, \theta_{t}) := \EE[\EE[Y \given X, Z, \Do(D=1, \Theta=\theta_t)]|\Do(D=1, \Theta = \theta_{t+1})] \\
				- \EE[Y \given \Do(D=1, \Theta = \theta_t)];
			\end{multline}
		\item[T3.] the baseline predictor $\EE[Y \given X, \Do(D=0)]$ is identifiable from $\{\PP(X, Y, Z \given \Do(D=1, \Theta=\theta)), \PP(X, Z \given \Do(D=0))\}$ via
			\begin{equation}\label{eqn:t3}
				\EE[Y \given X, \Do(D=0)]	= \EE[\EE[Y \given X, Z, \Do(D=1, \Theta=\theta)]|X, \Do(D=0)].
			\end{equation}
	\end{enumerate}
\end{corollary}

We note that the assumption in Proposition \ref{thm:id_complete} of availability of measurements of a domain pivot $\{X, Z\}$ from the target domain is necessary for solving T1--3. Indeed, Lemma \ref{thm:da_reduces_to_regression}, Proposition \ref{thm:unbiased_ce_not_identified} and Proposition \ref{thm:id_complete} together show the necessity of these measurements to solve these estimation tasks, if one is not willing to make further assumptions on the causal model.

\section{Estimation}\label{sec:estimation}
The identification result for the quantity $\EE[Y \given X, \Do(D, \Theta)]$ expresses the target quantity as an integral of a conditional expectation; this expression does not indicate how to \emph{estimate} the quantity of interest. When $X$ and $Z$ have finite sample spaces, one can estimate the conditional expectation (the integrand) with maximum likelihood, and compute the integral as a finite sum. However, when there are continuous variables involved, one has to tend to regression methods. In this section we expand on the suitable \emph{repeated regression} procedure, as proposed by \cite{boeken2023correcting}.

\medskip
To estimate $\EE[Y \given X, \Do(D=d, \Theta=\theta)]$ we can express equation (\ref{eqn:id_unbiased_model}) as
\begin{equation}\label{eqn:target_exp}
	\EE[Y \given X, \Do(D=d, \Theta=\theta)] = \EE[\EE[Y \given X, Z] \given X, \Do(D=d, \Theta=\theta)],
\end{equation}
where we used $Y\Indep D, \Theta \given X, Z$ to remove the conditioning on $(D, \Theta)$ in the inner expectation.\footnote{Formally, the independence $Y\Indep D, \Theta \given X, Z$ between random variable $Y$ and input variables $D, \Theta$ is to be interpreted as transitional conditional independence \citep{forre2021transitional}. It implies $\EE[Y|X, Z] = \EE[Y|X, Z, \Do(D=d, \Theta=\theta)]$~~$\PP(X, Z | \Do(D=d, \Theta=\theta))$-a.e., so we may pool data over multiple epochs when estimating $\EE[Y|X, Z]$.}
We formulate an estimation procedure based on this expression by estimating both conditional expectations on the right-hand-side with a regression model. More explicitly, given a sample $(X_i, Y_i, Z_i) \sim \PP(X, Y, Z \given \Do(D=d', \Theta=\theta'))$ with indices $i$ in index set $\Ical_S$ (source) and $(X_i, Z_i) \sim \PP(X, Z \given \Do(D=d, \Theta=\theta))$ with indices $i$ in index set $\Ical_T$ (target), the \emph{repeated regression estimator} $\hat{\EE}[Y \given X, \Do(D=d, \Theta=\theta)]$ is defined by estimating $\hat{\EE}[Y \given X, Z]$ from $\Ical_S$, augmenting the target dataset with pseudo-labels $\widetilde{Y}_i := \hat{\EE}[Y \given X=X_i, Z=Z_i]$ for all $i\in \Ical_T$, and estimating $\hat{\EE}[Y \given X, \Do(D=d, \Theta=\theta)] :=\hat{\EE}[\widetilde{Y} \given X]$ on the augmented target dataset.

For estimation of the expectation $\EE[Y|\Do(D=d, \Theta=\theta)]$, we can estimate a regression model $\hat{\EE}[Y \given X, Z]$ on the source dataset $\Ical_S$, and directly compute $\hat{\EE}[Y|\Do(D=d, \Theta=\theta)] := |\Ical_T|^{-1}\sum_{i \in \Ical_T}\hat{\EE}[Y \given X=x_i, Z=z_i]$.

The repeated regression procedure only requires measurements of the variable $Z$ to be available during training and not during deployment. Hence, this estimation procedure falls under the Learning using Privileged Information paradigm \cite{vapnik2009new,vapnik2015learning}. This is a convenient property, as it might be costly or even impossible to measure the covariates $Z$ at test time. We leave the choice of the regression method up to the practitioner, but we remark that these methods typically impose further assumptions on the sample spaces, exogenous distributions and structural equations.

\medskip
Using Example \ref{ex:naive_fails}, we demonstrate how these methods can be used to evaluate the deployment of the DSS.
\setcounter{example}{0}
\begin{example}[Application A: Evaluation]
	Recall the data generating process $X \sim \mathrm{Unif}[0,1], \hat{Y} = f(X)$ for some function $f$, but now with intermediary variables $A = D \cdot \I\{\hat{Y} > 1/2\}$ with $Y \sim \text{Ber}(\sigma(X-1/2)\cdot (1-A))$. Recalling Figure \ref{fig:ex1_naive}, we will compute $\tau$ or $\rho$ between the epochs to see whether deployment of a new model would be the right choice. Since $Y\Indep D, \Theta \given X, A$, we justifiably use $\{X, A\}$ as domain pivot for estimating $\tau$ and $\rho$. Between $t=1$ and $t=2$, for a model with parameter value $\theta_2$ to be deployed in epoch 2, we can sample $\PP(X, A | \Do(D=1, \Theta=\theta_2))$, and estimate $\tau(\theta_2) \approx -0.47$ using equation (\ref{eqn:t1a}), with the iterated expectation computed with repeated (polynomial logistic) regression. The lower the value of $Y$ the better, so we decide to deploy this model in epoch $t=2$, as displayed in Figure \ref{fig:ex1_naive:ep2}. Between epochs $t=2$ and $t=3$, we have retrained a model with parameter value $\theta_3$. Before deploying it, we can sample $\PP(X, A | \Do(D=1, \Theta=\theta_3))$, and estimate $\rho(\theta_3, \theta_2) \approx 0.47$ using equation (\ref{eqn:t2}), based on which we can decide not to deploy it.
\end{example}

Continuing with Example \ref{ex:naive_fails}, we demonstrate that the repeated regression procedure correctly estimates the baseline predictor $\EE[Y \given X, \Do(D=0)]$ from $\PP(X, A \given \Do(D=0))$ and $\PP(X, A, Y \given \Do(D=1, \Theta))$, and thereby provides a stable estimation procedure for retraining prediction models.

\begin{figure}[!htb]
	\centering
	\begin{tikzpicture}
		\node[inner sep=0pt] (b) at (0, 0) {\includegraphics[height=110pt]{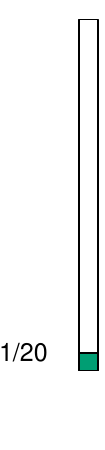}};
		\node[inner sep=0pt] (b) at (2.3, 0) {\includegraphics[height=110pt]{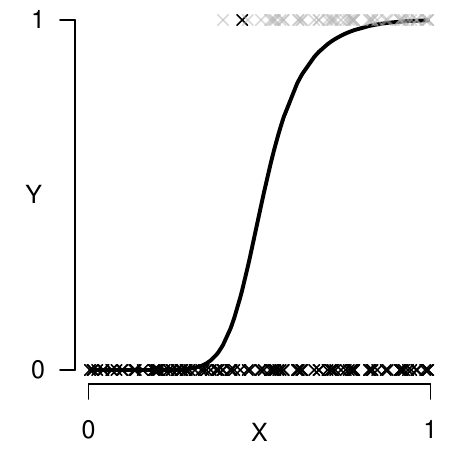}};
		\node[inner sep=0pt] (b) at (0.3, -1.73) {\scriptsize$\EE_3[Y]$};
		\node[inner sep=0pt] (b) at (2.1, -.6) {\scriptsize$\hat{Y}$};
		\draw[arr,draw=cbGreen] (3.3, 1.6) -- (3.3, -1.1);
		\draw[arr,draw=cbGreen] (3.7, 1.6) -- (3.7, -1.1);
	\end{tikzpicture}
	\caption{$t=3, \hat{Y} = \hat{\EE}[Y|X,\Do(D=0)]$ \label{fig:ex1_rr:ep4}}
	\label{fig:ex1_rr}
\end{figure}

\setcounter{example}{0}
\begin{example}[Application B: Bias correction]
	As mentioned above, a model with parameter value $\theta_3$ that is naively trained on data from epoch $t=2$ suffers from performative bias, as is found by estimating $\rho(\theta_3, \theta_2)$. Instead of deploying the naively retrained model, we leverage the domain pivot $\{X, A\}$ to estimate the baseline predictor $\hat{Y}_{bp} = \EE[Y|X, \Do(D=0)]$ using repeated (polynomial logistic) regression. This model is displayed in Figure \ref{fig:ex1_rr:ep4}.
\end{example}

\vspace{-10pt}
\section{Sample selection bias and selective labelling}\label{sec:sbias}
When dealing with real-world data, it is not uncommon that the data suffers from some form of \emph{sample selection bias}: that units are filtered before they are being measured, rendering the sample unrepresentative of the population.
Another, related form of bias is when units are selectively labelled, i.e.\ when the label $Y$ can be missing. This is explicitly prevalent in human-algorithmic decision making, when based on some prediction $\hat{Y}$ the unit can be dismissed, and the outcome $Y$ is not measured; see also \cite{guerdan2023ground}.
These selection mechanisms can be causally modelled by including a binary sample selection indicator $S^s$ and binary labelling indicator $S^\ell$ in the SCM, where $S^s=1$ indicates that the unit is included in the dataset, and $S^\ell=1$ indicates that the label $Y$ is observed.
\begin{figure}[!htb]
	\centering
	\begin{tikzpicture}
		\node[var] (X) at (0, 0) {$X$};
		\node[var] (Yh) at (1.5, 0) {$\hat{Y}$};
		\node[var] (A) at (3, 0) {$A$};
		\node[var] (C) at (3.75, 1.25) {$C$};
		\node[var] (Y) at (4.5, 0) {$Y$};
		\node[var] (M) at (4.5, -1.35) {$M$};
		\node[vari] (theta) at (0.75, 1.25) {$\Theta$};
		\node[vari] (D) at (2.25, 1.25) {$D$};
		\node[var] (S) at (3, -1.35) {$S$};
		\draw[arr] (X) to (Yh);
		\draw[arr] (M) to (Y);
		\draw[arr, bend right] (X) to (S);
		\draw[arr] (C) to (A);
		\draw[arr] (C) to (Y);
		\draw[arr, bend left] (C) to (S);
		\draw[arr, bend right] (X) to (A);
		\draw[arr, bend right] (X) to (Y);
		\draw[arr] (theta) to (Yh);
		\draw[arr] (Yh) to (A);
		\draw[arr] (D) to (A);
		\draw[arr] (S) to (M);
		\draw[arr] (A) to (S);
		\draw[arr] (A) to (Y);
	\end{tikzpicture}
	\caption{A causal graph with selection variable $S$.}
	\label{fig:selection_combined}
\end{figure}
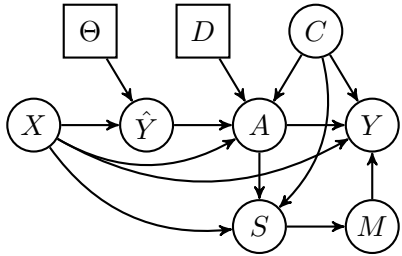

For ease of notation, we let $S = S^s \wedge S^\ell$, so that in the source domain we measure data from $\PP(X, Y, Z \given S=1, \Do(D, \Theta))$, with $Z\subseteq V$ a set of variables. If one wants to correct for selection bias, the target quantity becomes $\EE[Y \given X, \Do(D=d, \Theta=\theta)]$, so without the conditioning on $S=1$. Similar to the treatment in Section \ref{sec:domain_adaptation} it can be shown that this target quantity is not identifiable, but one can extend Proposition \ref{thm:id_complete} to settings where selection bias is in play by considering $(D, \Theta, S)$ to be the domain indicator and $\{X, Z\}$ a domain pivot, so with $Y\Indep D,\Theta, S \given X, Z$. For example, in Figure \ref{fig:selection_combined} we can let $Z = \{A, C, M\}$, where $A$ can for example be an action, $C$ a confounder, and $M$ a mediator of the selection variable and the outcome.
The target quantity $\EE[Y \given X, \Do(D=d, \Theta=\theta)]$ is then identifiable in $\Mcal$ from $\{\PP(X, Y, Z \given S=1, \Do(D=d', \Theta=\theta')), \PP(X, Z \given \Do(D=d, \Theta=\theta))\}$ if also $\PP(X, Z \given \Do(D=d, \Theta=\theta)) \ll \PP(X, Z \given S=1, \Do(D=d', \Theta=\theta'))$, in which case
\begin{equation}\label{eqn:id_unbiased_model_sb}
	\EE[Y \given X, \Do(D=d, \Theta=\theta)] = \EE[\EE[Y \given X, Z, S=1]\given X, \Do(D=d, \Theta=\theta)]
\end{equation}
$\PP(X|\Do(D=d, \Theta=\theta))$-a.e. This iterated expectation can be estimated using repeated regression.

A similar identification result has been given in \cite{boeken2023correcting}, but solely for selection bias or missing response. The above identification result shows the ability of repeated regression to correct for multiple forms of domain shift simultaneously, provided that a domain pivot can be measured in the target domain.
For more information on selection bias and missing response, we refer to \cite{boeken2023correcting} and references therein.

\section{Discussion}
In this work, we modelled the deployment of a decision support system as causal domain shift, introduced evaluation and bias correction as two applications of this causal model, and have shown how certain estimands in these applications can be only be estimated under the availability of measurements of a \emph{domain pivot} in the target domain. We have demonstrated how \emph{repeated regression} is a suitable estimation procedure for evaluation and bias correction, even if the measured labels are subject to selection bias and/or selective labelling.

Sensitivity analysis with respect to the conditional independence assumption, such as the estimation of bounds of the quantity of interest when this independence does not hold, might be a promising direction for future work. Constructing doubly robust and efficient estimators for the deployment effect, retraining effect, and baseline predictor, e.g.\ using influence functions \citep{dudik2014doubly,athey2019surrogate}, would also be of interest.

An important assumption for the relevance of our identifiability results is that labels are never observed in the target domain of the domain adaptation problems. Estimating the deployment effect, retraining effect, and baseline predictor can be done directly on available data if labels are measured in the target domain, e.g.\ through A/B testing of the deployment of the DSS. However, in high stakes environments it is often neither desired nor ethical to carry out such experiments. If these labels are not measured in the target domain, our proposed methods heavily depend on the availability of measurements of a domain pivot in the target domain, without actually deploying the DSS. This can be a restrictive assumption in practice. If such data is available, our evaluation metrics rely on an estimated model; a practice that requires caution.

Nevertheless, we hope that the proposed causal model, evaluation metrics and the concept of performative bias will be useful tools for responsible applications of AI systems in high-stakes settings.

\acks{%
	We thank the anonymous reviewers, Stephan Bongers and Sourbh Bhadane for their feedback, which considerably helped improving the manuscript. This work is supported by Booking.com.
}

\bibliography{refs}

\appendix

\section{Relation to existing literature}
\subsection{Performative prediction}\label{app:perf_pred}
Translated to our notation, \cite{perdomo2020performative} introduced the performative risk:
\begin{equation}
	\EE[\ell(\hat{Y}, Y) | \Do(D=1, \Theta=\theta)].
\end{equation}
where $\ell$ is a loss function. They define a parameter $\theta_t$ to be performatively stable if it constant under retraining, so if we get
\begin{equation}
	\theta_t \approx \argmin_\theta\EE[\ell(\hat{Y}, Y) | \Do(D=1, \Theta=\theta_t)].
\end{equation}
They don't consider the case where $D=0$, so the deployment effect $\tau(\theta)$ can for example not be defined using the existing framework.

The baseline predictor $\hat{Y} = \EE[Y|X, \Do(D=0)]$ does not minimize performative risk, but in the setting of Proposition \ref{thm:baseline_predictor_optimal}, it is a performatively stable predictor:
if we parametrise $\hat{Y}_{\theta_t} = \hat{\EE}_{\theta_t}[Y|X, \Do(D=0)]$, then estimating $\hat{Y}_{\theta_{t+1}} = \hat{\EE}_{\theta_{t+1}}[Y|X, \Do(D=0)]$ from $\PP(X, Y, Z | \Do(D=1, \Theta=\theta_t))$ and $\PP(X, Z | \Do(D=0))$ will yield $\theta_t \approx \theta_{t+1}$. A performatively stable parameter $\theta_t \approx \theta_{t+1}$ will have as retraining effect $\rho(\theta_t, \theta_{t+1}) \approx 0$.

\subsection{Off-policy evaluation}\label{app:off-policy}
In contextual bandits, one considers a context $X$, an action $A\sim\PP(A|X, \Do(\Theta=\theta))$ (where $\PP(A|X, \Do(\Theta))$ is referred to as a \emph{policy} with parameters $\Theta$), and a reward $Y \sim \PP(Y|X, A)$. This gives rise to a joint distribution $\PP(X, A, Y | \Do(\Theta))$. When one has measured data from a policy with parameters $\theta$, the problem of \emph{off-policy evaluation} is that of estimating for a new set of parameters $\theta'\neq \theta$ the reward $\EE[Y|\Do(\Theta=\theta')]$ from $\PP(X, A, Y | \Do(\Theta=\theta))$.

If the relation between $X, A$ and $Y$ is such as described above, Proposition \ref{thm:id_complete} combined with the repeated regression procedure says that we can estimate
\begin{equation}
	\EE[Y|\Do(\Theta=\theta')] = \EE[\EE[Y|X, A] | \Do(\Theta=\theta')],
\end{equation}
which is known as the \emph{direct method} in contextual bandit literature \cite{dudik2014doubly}.
Note that our results from Section \ref{sec:domain_adaptation} consider a rather intricate policy, namely one that factorizes according to $\PP(A | X, C, \Do(D=1, \Theta = \theta)) = \PP(A | X, C, \hat{Y}, \Do(D=1))\PP(\hat{Y} | X, \Do(\Theta=\theta))$. Note that typically, for given parameters $\theta$, we don't \emph{know} the policy $\PP(A | X, C, \Do(D=1, \Theta = \theta))$ (contrary to when one considers a setting of automated decision making) but we can merely sample from it, as is explained in Section \ref{sec:domain_adaptation:domain_pivot}.

\subsection{Surrogate indices}\label{app:surrogates}
\cite{athey2019surrogate} consider the estimation of a causal effect with a similar technique as repeated regression. For estimating a causal effect
\begin{equation}
	\EE[Y|\Do(D=1)] - \EE[Y|\Do(D=0)],
\end{equation}
they consider the case where one has two samples: an observational sample with measurements of covariates $X$, target variable $Y$ and so-called \emph{surrogates} $Z$ (so not of $D$), and an experimental sample with measurements of $X, D$ and $Z$ (so not of $Y$). For a set of variables $Z$ to be surrogates, they require the independence $Y\Indep D \given X, Z$\footnote{This is the same conditional independence that we require for the domain pivot, but we are reluctant to call a domain pivot a surrogate, as we don't restrict the domain shift to be the value of an intervention but also other types of domain shift, like selection bias.}, and that $\{X, Z\}$ is a valid \emph{adjustment set} for estimating the causal effect of $D$ on $Y$, i.e.\ $\EE[Y|\Do(D)] = \EE[\EE[Y|X, Z] | D]$. Their identification strategy is built on the equation
\begin{equation}
	\EE[Y|\Do(D=1)] - \EE[Y|\Do(D=0)] = \EE[\EE[Y|X, Z] | D=1] - \EE[\EE[Y|X, Z] | D=0].
\end{equation}

Note that we consider a different setup. Instead of having `observational' and `experimental' samples, alternating the measurements of treatment $D$ or outcome $Y$, we consider a setting where for one value of the treatment ($D=0$, say) we observe $Y$, and for the other value of the treatment we don't observe $Y$. An overview of these different assumptions is given in Table \ref{tab:compare_athey}. Our identification result is similar to that of \cite{athey2019surrogate}, but can be interpreted as intervention extrapolation, instead of causal effect estimation using surrogate outcomes.

\begin{table}[htb]
	\centering
	\begin{tabular}{lllllll}\toprule
		                          & \textbf{Sample} & $X$        & $D=0$      & $D=1$      & $Z$        & $Y$        \\
		\midrule
		\cite{athey2019surrogate} & Observational   & \checkmark & $\times$   & $\times$   & \checkmark & \checkmark \\
		                          & Experiment      & \checkmark & \checkmark & \checkmark & \checkmark & $\times$   \\
		\midrule
		This work                 & Source ($D=0$)  & \checkmark & \checkmark & $\times$   & \checkmark & \checkmark \\
		                          & Target ($D=1$)  & \checkmark & $\times$   & \checkmark & \checkmark & $\times$   \\
		\bottomrule
	\end{tabular}
	\caption{A comparison of the setting in \cite{athey2019surrogate} and this work.}
	\label{tab:compare_athey}
\end{table}

\section{Proofs}
\begin{proof}[Proposition \ref{thm:baseline_predictor_optimal}]
	Define $G := \{x : \PP(Y=1 \given X=x, A=0) > \varepsilon(x)\}$. We have the unique optimal policy $A^*(X) := \I_G(X)$ for $\min_{a \in \{0,1\}^\Xcal} \PP(Y=1 \given X, A=a)$, and thus we have the set of minimizers
	\begin{align*}
		H & := \argmin_{\hat{y} \in [0,1]^\Xcal}\PP(Y=1 \given X=x, \Do(\hat{Y} = \hat{y}(x)))          \\
		  & = \argmin_{\hat{y} \in [0,1]^\Xcal}\PP(Y=1 \given X=x, A = \I\{\hat{y}(x) > \varepsilon\})  \\
		  & = \{\hat{y}\in[0,1]^\Xcal\ : A^*(x) = \I\{\hat{y}(x) > \varepsilon(x)\} \forall x\in\Xcal\} \\
		  & = \{\hat{y}\in[0,1]^\Xcal\ : \hat{y}(x) \geq \PP(Y=1 \given X=x, A=0)\}.
	\end{align*}
	Clearly we have
	\begin{align*}
		\hat{Y}^* & := \argmin_{\hat{y}\in H} \PP(A=1 \given X=x, \Do(\hat{Y} = \hat{y}(x))) \\
		          & = \argmin_{\hat{y}\in H} \I\{\hat{y}(x)\geq \varepsilon(x)\}             \\
		          & = \PP(Y=1 \given X=x, A=0),
	\end{align*}
	and since $Y\Indep D \given X, A$ and $D=0 \implies A=0$ we further get
	\begin{equation*}
		\PP(Y=1 \given X=x, A=0) = \PP(Y=1 \given X=x, A=0, D=0) = \PP(Y=1 \given X=x, D=0),
	\end{equation*}
	and thus $\hat{Y}^* = \EE[Y \given X, D=0]$.
\end{proof}

\begin{proof}[Lemma \ref{thm:da_reduces_to_regression}]
	If $\EE[Y \given X, \Do(D=d, \Theta=\theta)]$ is identifiable from $\PP(X, Y \given \Do(D = d', \Theta = \theta'))$, then since $X\Indep D, \Theta$ we have $\PP(X \given \Do(D=d', \Theta=\theta')) = \PP(X \given \Do(D=d, \Theta=\theta))$, so $\EE[Y|\Do(D=d, \Theta=\theta)] = \EE[\EE[Y \given X, \Do(D=d, \Theta=\theta)]\given \Do(D=d', \Theta=\theta')]$, so $\EE[Y|\Do(D=d, \Theta=\theta)]$ is identifiable as well.

	If $\EE[Y \given X, \Do(D=d, \Theta=\theta)]$ is not identifiable from $\PP(X, Y \given \Do(D = d', \Theta = \theta'))$, then there exist $M_1$ and $M_2$ such that $\PP_{M_1}(X, Y \given \Do(D = d', \Theta = \theta')) = \PP_{M_2}(X, Y \given \Do(D = d', \Theta = \theta'))$ and $\EE_{M_1}[Y \given X, \Do(D=d, \Theta=\theta)] \neq \EE_{M_2}[Y \given X, \Do(D=d, \Theta=\theta)]$. Let $x'$ be such that $\EE_{M_1}[Y \given X=x', \Do(D=d, \Theta=\theta)] \neq \EE_{M_2}[Y \given X=x', \Do(D=d, \Theta=\theta)]$ and let $\tilde{M}_1, \tilde{M}_2$ be equal to the SCMs $M_1, M_2$, except for the structural equation for $X$, which is set to $X = x'$ in both $\tilde{M}_1$ and $\tilde{M}_2$. Then we still have $\PP_{\tilde{M}_1}(X, Y \given \Do(D = d', \Theta = \theta')) = \PP_{\tilde{M}_2}(X, Y \given \Do(D = d', \Theta = \theta'))$, and $\EE_{\tilde{M}_1}[Y|\Do(D=d, \Theta=\theta)] = \EE_{\tilde{M}_1}[Y \given X=x', \Do(D=d, \Theta=\theta)] \neq \EE_{\tilde{M}_2}[Y \given X=x', \Do(D=d, \Theta=\theta)] = \EE_{\tilde{M}_2}[Y|\Do(D=d, \Theta=\theta)]$, so $\EE[Y \given \Do(D=d, \Theta=\theta)]$ is not identifiable from $\PP(X, Y \given \Do(D = d', \Theta = \theta'))$.
\end{proof}

\begin{proof}[Proposition \ref{thm:unbiased_ce_not_identified}]
	Let $\theta, \theta' \in \RR, d, d' \in \{0,1\}$ be given, and consider for $i=1, 2$ the SCM $M_i$ given by $X\sim \Ncal(0,1), \hat{Y} = \Theta \cdot X + i\cdot \I\{\Theta\neq \theta'\}, Y = X + \I\{D=1\}\cdot \hat{Y} + i\cdot \I\{D\neq d'\}$.
	One can readily verify that $\PP_{M_1}(X, Y \given \Do(D=d', \Theta=\theta')) = \PP_{M_2}(X, Y \given \Do(D=d', \Theta=\theta'))$, but that $\EE_{M_1}[Y \given X, \Do(D=d, \Theta=\theta)] = 1 + X \neq 2 + X = \EE_{M_2}[Y \given X, \Do(D=d, \Theta=\theta)]$ if $d=0$, that $\EE_{M_1}[Y \given X, \Do(D=d, \Theta=\theta)] = (\theta + 1) X + 1 \neq (\theta + 1) X + 2 = \EE_{M_2}[Y \given X, \Do(D=d, \Theta=\theta)]$ if $d=1$ and either $d=d'$ or $\theta = \theta'$, and that $\EE_{M_1}[Y \given X, \Do(D=d, \Theta=\theta)] = (\theta + 1) X + 2 \neq (\theta + 1) X + 4 = \EE_{M_2}[Y \given X, \Do(D=d, \Theta=\theta)]$ if $d=1$ and both $d\neq d'$ and $\theta \neq \theta'$.
\end{proof}

\begin{proof}[Proposition \ref{thm:id_complete}]
	That $Y\Indep D, \Theta \given X, Z$ implies identifiability (under the required positivity assumption) is immediate from equation (\ref{eqn:id_unbiased_model}).

	If $Y\nIndep D\given X, Z$, we note that for all $Z' \in Z$, we cannot have $X \to Z' \to \hat{Y}$ as this violates Assumption \ref{ass:pp_graph}. If there is an edge $\hat{Y} \to Y$, we can augment the constructed $M_1, M_2$ from the proof of Proposition \ref{thm:unbiased_ce_not_identified} where we let all $Z' \in Z$ be independent variables having the same distribution in both models, which proves non-identifiability in that setting. The last case to check is where every directed path from $\hat{Y}$ to $Y$ contains at least one element from $Z$. Since we have $Y\nPerp^d_{G'} D \given X, Z$, there is at least one such a path $\pi = \hat{Y} \to Z_1 \to ... \to Z_n \to Y$ for some $n\in\NN$ with $Z_1, ..., Z_n \in Z$ and $Z_1\leftrightarrow Y$ in $G'$. We define $M_1$ by letting $X \sim \Ber(1/2), \hat{Y}=X, E_{Z_1} \sim\Ber(1/2), E_{Z_1Y} \sim \Ber(1/2), Z_1 = \I_{\{D\neq d\}}\cdot\XOR(D, E_{Z_1Y}) + \I_{\{D=d\}}\cdot\XOR(D, E_{Z_1}), Z_{i+1} = Z_i$ for $i=1, ..., n-1$, and $Y = \XOR(Z_n, E_{Z_1Y})$. We let all other variables in $Z$ be independent. Define $M_2$ to be equal to $M_1$, with the only difference that $Z_1 = \XOR(D, E_{Z_1Y})$. Then indeed $\PP_1(X, Y, Z \given \Do(D=d', \Theta))= \PP_2(X, Y, Z \given \Do(D=d', \Theta))$, $\PP_1(X, Z \given \Do(D=d, \Theta)) = \PP_2(X, Z \given \Do(D=d, \Theta))$ and $\EE_1[Y \given X, \Do(D=d, \Theta)] = d \neq 1/2 = \EE_2[Y \given X, \Do(D=d, \Theta)]$, proving non-identifiability.
\end{proof}

\end{document}